\documentclass[letterpaper, 10 pt, conference]{ieeeconf}  

\IEEEoverridecommandlockouts                              

\newif\ifmargincomments 
\margincommentstrue

\usepackage{graphicx} 
\usepackage{epsfig} 
\usepackage{amsmath} 
\usepackage{amssymb}  

\usepackage{amsthm}
\usepackage{mathtools}
\usepackage{algorithm}
\usepackage[noend]{algpseudocode}
\usepackage{flushend}

\usepackage[usenames,dvipsnames]{xcolor}

\ifmargincomments

\else

\fi

\newtheorem{theorem}{Theorem}

\newcommand{\pushright}[1]{\ifmeasuring@#1\else\omit\hfill$\displaystyle#1$\fi\ignorespaces}

\graphicspath{ {fig/} }

\title{\LARGE \bf
Data-Driven Model Predictive Control of \\Autonomous Mobility-on-Demand Systems
}

\author{Ramon Iglesias$^{1}$ \quad Federico Rossi$^{2}$ \quad Kevin Wang$^{3}$ \quad David Hallac$^{4}$\quad Jure Leskovec$^{5}$ \quad Marco Pavone$^{2}$
\thanks{$^{1}$Ramon Iglesias is with the Department of Civil and Environmental Engineering,
        Stanford University, Stanford, CA, 94305,
       {\tt\small rdit@stanford.edu}.}%
\thanks{$^{2}$Federico Rossi and Marco Pavone are with the Department of Aeronautics and Astronautics, Stanford University, Stanford, CA, 94305,
       \{{\tt\small frossi2, pavone}\} {\tt \small @stanford.edu}.}%
\thanks{$^{3}$Kevin Wang is with Houzz Inc.,
        Palo Alto, CA 94301,
       {\tt\small kevin.wang@houzz.com}.}%
\thanks{$^{4}$David Hallac is with the Department of Electrical Engineering,
        Stanford University, Stanford, CA, 94305,
       {\tt \small hallac@stanford.edu}.}%
\thanks{$^{5}$Jure Leskovec is with the Department of Computer Science,
        Stanford University, Stanford, CA, 94305,
       {\tt \small jure@stanford.edu}.}%
\thanks{This research was supported by the National Science Foundation under CAREER Award CMMI-1454737 and by the Toyota Research Institute (``TRI"). This article solely reflects the opinions and conclusions of its authors and not NSF, TRI, or any other Toyota entity.}
}

\begin{document}

\maketitle
\thispagestyle{empty}
\pagestyle{empty}

\begin{abstract}

The goal of this paper is to present an end-to-end, data-driven framework to control Autonomous Mobility-on-Demand systems (AMoD, i.e. fleets of self-driving vehicles). 
We first model the AMoD system using a time-expanded network, and present a formulation that computes the optimal rebalancing strategy (i.e., preemptive repositioning) and the minimum feasible fleet size for a given travel demand.
Then, we adapt this formulation to devise a Model Predictive Control (MPC) algorithm that leverages short-term demand forecasts based on historical data to compute rebalancing strategies. Using simulations based on real customer data from DiDi Chuxing, we test the end-to-end performance of this controller with a state-of-the-art LSTM neural network to predict customer demand: we show that this approach scales very well for large systems (indeed, the computational complexity of the MPC algorithm does not depend on the number of customers and of vehicles in the system) and outperforms state-of-the-art rebalancing strategies by reducing the mean customer wait time by up to to 89.6 \%.

\end{abstract}

\section{Introduction}
The last decade has seen both the emergence and rapid expansion of Mobility-on-Demand (MoD) services (e.g. carsharing and ridesharing) and the promising development of self-driving technology. 
Together, MoD services and self-driving vehicles provide the key components to a new form of transportation called Autonomous Mobility-on-Demand (AMoD), wherein a fleet of self-driving vehicles provides travel services on-demand. 

A key operational challenge for AMoD systems, as for any transportation system with asymmetric demand, is the problem of {\emph{imbalance}}:  vehicles naturally concentrate in a subset of the areas serviced by the MoD system, limiting the availability in other regions \cite{Fricker2012,David2012}. 

Devising efficient operating strategies for the imbalance problem is an active area of research for MoD and AMoD systems. However, the majority of the existing body of work does not leverage the ability to forecast customer demand.
Accordingly, most existing control strategies are \emph{reactive}: thus, they do not deal well with rapidly time-varying demand due to, e.g., commuting cycles, events, or weather phenomena. 
 The goal of this paper is to present an end-to-end, data-driven framework to control AMoD systems with a focus on the imbalance problem: by leveraging information about predicted customer demand in the control synthesis problem, we design a \emph{predictive} control strategy that anticipates imbalances in customer demand and rebalance vehicles accordingly, and we demonstrate the performance of such strategy with real-world data.




{\emph{Literature Review.}} A considerable amount of research has been devoted to the design and analysis of optimal control of taxi-like fleets. 
In non-autonomous MoD systems such as Uber, and Lyft, ridesharing operators often use pricing incentives, commonly known as dynamic pricing, to nudge drivers towards areas where demand outstrips supply. 
In \cite{Banerjee2016} the authors provide a framework for synthesizing pricing policies; \cite{Banerjee2015} shows that dynamic pricing is not necessarily better than the optimal static policy, but is more robust with respect to system parameter uncertainties. 

However, in AMoD systems, the fleet operator can \emph{directly} control the routes and schedules of the autonomous vehicles. \cite{PavoneSmithEtAl2012,ZhangPavone2016,ZhangRossiEtAl2016,VolkovAslamEtAl2012} approach the problem of controlling a fleet by first formulating steady-state solutions using queuing theoretical \cite{ZhangPavone2016}, fluidic \cite{PavoneSmithEtAl2012}, network flow \cite{ZhangRossiEtAl2016}, or Markov \cite{VolkovAslamEtAl2012} models, respectively, and then deriving heuristic control laws informed by the steady-state solution that can be applied in real-time.
However, by relying on steady-state formulations, the aforementioned pricing or control heuristics are time-invariant, and, in particular, cannot accommodate time-varying forecasted demand. 
Time-varying, MPC controllers have been proposed in \cite{ZhangRossiEtAl2016b,Miao2016}, and the methods in \cite{Miao2016,Miller2017} explicitly consider forecasted demand. 
In particular, the Model Predictive Control (MPC) approach from \cite{ZhangRossiEtAl2016b} experimentally outperforms the time-invariant heuristics listed above, and could potentially leverage forecasted demand. However, by assigning decision variables to each vehicle in \cite{ZhangRossiEtAl2016b,Miao2016} and enumerating all positioning possibilities in \cite{Miller2017}, the problem sizes in these methods grow significantly for large fleets, which limits their real-life applications to small or medium fleets. Moreover, the model in \cite{Miao2016} requires that vehicles should be able to pickup and dropoff customers within a time step, limiting its applicability to systems where all customer travel times are similar.

{\emph{Statement of Contributions.}} Our contribution in this paper is threefold. First, we propose an efficient  approach to find the optimal dispatching policy for the case when the trip demand is \emph{known} ahead of time. This provides an upper bound on the performance of the system. The approach is able to simultaneously optimize the dispatching policy and the number of required vehicles: thus, it can be used for \emph{fleet sizing}. Second, we propose an MPC algorithm for operating the system in real-time by leveraging short-term forecasts of customer demand. The complexity of the algorithm does not depend on the number of vehicles or on the number of customers in the transportation system: thus, the algorithm can be used to effectively control large-scale AMoD systems. Third, we validate these approaches using a dataset of DiDi Chuxing, the major ridesharing company in China: our results show that the proposed MPC algorithm outperforms a state-of-the-art algorithm with a 89.6\% reduction in mean customer wait time.

{\emph{Organization.}} The rest of this paper is organized as follows: We first present in Section \ref{sec:model_description} a time-varying model for AMoD systems. In Section \ref{sec:problem_formulation}, we leverage the time-varying model to propose a MPC algorithm that relies on predicted future customer demand to control an AMoD system. 
Finally, in Section \ref{sec:numerical_experiments}, we validate the approach on a real-world scenario based on a dataset of DiDi Chuxing, and characterize the performance of the proposed controller as a function of the prediction quality. 

\section{Model Description and Problem Formulation}\label{sec:model_description}
In this section, we propose a time-varying network-flow model for AMoD systems that assumes perfect information is available about future customer arrivals. The model is amenable to efficient optimization: thus, it can be used to optimize the scheduling and routing for an AMoD system a posteriori. The model is not causal, and therefore can not directly be used for real-time control of an AMoD system; however, it forms the core of the MPC controller presented in the next section.

We consider an urban environment discretized into a set $\mathcal{N}$ of distinct regions (also known as \emph{stations} in the AMoD literature \cite{PavoneSmithEtAl2012,ZhangPavone2016}).  Time is represented by discrete intervals of a given size $\Delta t$. For a period under consideration of length $T$ time intervals, denote $\mathcal{T} = [1,...,T]$ as the ordered set of time intervals. An AMoD system provides transportation services. We denote the travel time experienced by self-driving vehicles traveling from region $i \in \mathcal{N}$ to region $j \in \mathcal{N}$ as $\tau_{ij}$: that is, a vehicle departing region $i$ at time $t$ arrives in region $j$ at time $t' = t + \tau_{ij}$. The travel time can change depending on the departure time (modeling time-varying traffic conditions); however, the travel time is assumed to be independent of the number of vehicles traveling between the regions (i.e., congestion is considered an exogenous phenomenon). 

Self-driving vehicles service customers' transportation requests. We denote the number of customers who wish to travel from region $i \in \mathcal{N}$ to another region $j \in \mathcal{N}$, departing at time $t$, as $\lambda_{ijt}$.
Similarly, we denote the number of vehicles that transport customers from $i \in \mathcal{N}$ to $j \in \mathcal{N}$ departing at time $t$ as  $x_{ijt}^p$.

In order to satisfy the customer demand at each given region and time interval, the number of available vehicles at each time and in each region must be no smaller than the number of customers who wish to depart the region. 
To this end, it is necessary to recurrently rebalance the empty vehicles from stations with an excess number of vehicles to stations with an insufficient number of available vehicles. We represent the number of empty, rebalancing vehicles traveling from $i \in \mathcal{N}$ to $j \in \mathcal{N}$ at time $t \in \mathcal{T}$ as $x_{ijt}^r$. When $i = j$, the vehicles are considered to be idling, and this is regarded as a special case of rebalancing.

Finally, let $s_{it}$ represent the number of initial available vehicles at region $i \in \mathcal{N}$ at time $t \in \mathcal{T}$. The variable $s_{it}$ is free: the optimizer is allowed to determine the number and location of vehicles that is required to service all customers. Initial available vehicles can only be added at the first time interval: $s_{it}=0\,\forall t>1$. The overall number of vehicles in the AMoD system is $m=\sum_{i\in\mathcal{N},t\in\mathcal{T}} s_{it}$.

\subsection{Optimal Rebalancing Strategy}\label{sec:static_problem}

Under the assumption of perfect knowledge of customer demand, and assuming that the starting positions of the vehicles are free, it is possible to find the optimal rebalancing strategy by solving the following optimization problem:

{\small
\begin{subequations}
\begin{align}
& \underset{\mathcal{X}^p, \mathcal{X}^r, \mathcal{S}}{\text{minimize}} 
  & & \sum_{(ijt)} c_{i,j}^r x_{ijt}^r,\\
  & \text{subject to} && x_{ijt}^p = \lambda_{ijt}, \quad \forall i,j \in \mathcal{N}, t \in \mathcal{T},\label{eq:pax-constr}\\
  &
  && \sum_{j \in \mathcal{N}} x_{ijt}^p + x_{ijt}^r - x_{jit-\tau_{ij}}^p - x_{jit-\tau_{ij}}^r = s_{it}, \nonumber\\
  &
  && \quad \quad \forall i \in \mathcal{N}, t \in \mathcal{T}\,\label{eq:veh-constr}\\ 
  &
  && s_{it} = 0, \quad \forall i\in \mathcal{N}, t \in \mathcal{T}, t > 1\,,\label{eq:start-constr}\\
  &
  && x_{ijt}^r, s_{i1} \in \mathbb{N}, \quad \forall i,j \in \mathcal{N}, t \in \mathcal{T}\,, \label{eq:int-vars}
\end{align}
\label{eq:det-opt}
\end{subequations}
} 
\noindent where $c_{ij}^r$ is the cost of rebalancing a vehicle from $i \in \mathcal{N}$ to $j \in \mathcal{N}$ (proportional to the travel time and distance between the regions). 
Idling vehicles are considered a special case of rebalancing denoted as $x_{iit}^r$ and have a corresponding cost $c_{iit}^r$, which captures time-only dependent costs (e.g. average cost of ownership per time interval) or costs specific to idling (e.g. parking). The decision variables are grouped into the following sets: $\mathcal{X}^p = \{x_{ijt}^p\}_{ijt}$, $\mathcal{X}^r = \{x_{ijt}^r\}_{ijt}$, $\mathcal{S} = \{s_{it}\}_{it}$.
The  constraint \eqref{eq:pax-constr} ensures that all customer demands are serviced, and constraint \eqref{eq:veh-constr} enforces that, for every time interval and each region, the number of arriving vehicles equals the number of departing vehicles. The constraint \eqref{eq:start-constr} ensures that starting vehicles can only be inserted at the first time interval, and \eqref{eq:int-vars} constrains the decision variables to be nonnegative integers. 

Problem \eqref{eq:det-opt} always admits a feasible solution. The next theorem formalizes this intuition.
\begin{theorem}[Feasibility of the offline optimal rebalancing problem] \label{th:feas}
Problem \eqref{eq:det-opt} admits a feasible solution for any set of customer demands $\{\lambda_{ijt}\}_{i,j,t}$.
\end{theorem}
\emph{Proof sketch:} The number and location of available vehicles is a decision variable in Problem \eqref{eq:det-opt}. Thus, a feasible solution exists where a vehicle is assigned to the departure location of each customer and idles at the location until the customer's departure time.

A few comments are in order. 
First, while Problem \eqref{eq:det-opt} is stated as an integer linear program (ILP), it can be shown that the problem is totally unimodular and thus can be solved efficiently as a linear program (LP) \cite{AhujaMagnantiEtAl1993}. Thus, very large instances of the problem can be solved efficiently on commodity hardware.
Second, note that while we model idling as a special case of rebalancing, we can give it a different treatment by relying on its cost parameter. For example, the downtown area of a city might impose a charge for the time spent within the district or parking might be expensive in certain areas, thus, incentivizing vehicles to move. In Section \ref{sec:numerical_experiments}, we assume that the cost per time interval of idling is simply related to time average cost of ownership, and, therefore, it is lower than rebalancing (since rebalancing additionally incurs, for example, fuel expenses).
Finally, the starting positions are decision variables: thus, Problem \eqref{eq:det-opt} simultaneously finds the optimal rebalancing strategy and the corresponding fleet size required to execute the strategy. Therefore, the solution to this problem can be used to inform fleet sizing decisions based on historical customer demand.

\section{Model Predictive Control}\label{sec:problem_formulation}

In this section, we propose an MPC implementation of the optimal rebalancing problem presented in Section \ref{sec:static_problem} that leverages predictions of future demand.
We begin by outlying the overall algorithm and then we delve into the details of each subcomponent.


\subsection{Algorithm}
Let $T$ be the planning horizon under consideration and $T_{\mathrm{forward}}$ the forecasting horizon for which we can obtain a predicted customer demand (note that $T_{\mathrm{forward}} \leq T$). Also, denote $\Lambda_{t,t'} = \{\lambda_{ijt} : \forall i,j \in \mathcal{N}, t \in [t,...,t']\}$ as the set of {\emph{real}} customer demands from $t$ to $t'$, and $\hat{\Lambda}_{t,t'} = \{\hat{\lambda}_{ijt} : \forall i,j \in \mathcal{N}, t \in [t,...,t']\}$ as the set of {\emph{predicted}} customer demands from $t$ to $t'$. $\{\lambda_{ij0}\}_{ij0}$ is the set of outstanding customer demand (customers who have requested a vehicle, but have not yet been served), and the rest of the notation is defined as in the previous section.

The proposed algorithm, summarized in Algorithm \ref{alg:mpc}, is as follows: at a given time $t_0$ we first observe the system state to capture the vehicle availabilities, $\mathcal{S}$, and the outstanding customer demand, $\{\lambda_{ij0}\}_{ij0}$. We then proceed to predict future customer demand $\hat{\Lambda}_{t_0,t_0+T_{\mathrm{forward}}}$ for the next $T_{\mathrm{forward}}$ time steps. Using this information, we compute the optimal rebalancing strategy $\mathcal{X}^r$ by solving a mixed-integer linear program (described in Section \ref{sec:controller}). Finally, we assign the rebalancing tasks corresponding to the {\emph{first}} time interval to available vehicles as they become available. 

After a period $\Delta t$, i.e. at $t_0 + 1$, we recompute the rebalancing strategy using Algorithm \ref{alg:mpc}. Thus, this process is repeated during the entire operation of the system.

\begin{algorithm}
\caption{Model Predictive Control}\label{alg:mpc}
\begin{algorithmic}[1]
\Procedure{MPC}{}
\State $\mathcal{S} \gets \text{count idle vehicles and estimate trip arrivals}$
\State $\lambda_{ij0} \gets \text{count outstanding customers}$
\State $\hat{\Lambda}_{t_0,t_0+T_{\mathrm{forward}}} \gets f(\theta_t)$
\State $\mathcal{X}^p, \mathcal{X}^r, \mathcal{W},\mathcal{D} \gets \text{solve Problem \eqref{eq:mpc-opt}}$
\State Assign $\{x_{ij1}^r\}_{ij1}$ to available vehicles
\EndProcedure
\end{algorithmic}
\end{algorithm}

\subsection{State observation}
At the beginning of each iteration of Algorithm \ref{alg:mpc}, the first step is to capture the current system state in terms of vehicle availabilities and outstanding passengers.

Unlike in Section \ref{sec:static_problem}, where the vehicles' starting locations, $\mathcal{S}$, are decision variables, during the operation of the system, the time and location at which the vehicles become available are fully determined by the current system state. Vehicles are considered available either when they are idling or after they complete a trip. 
Thus, at the start of the optimization process, let $a_{i}$ be the current number of idling vehicles at region $i \in \mathcal{N}$. Additionally, let $v_{it}$ be the number of vehicles traveling to region $j \in \mathcal{N}$ expected to arrive at time interval $t$. Then, the vehicle starting locations for the planning horizon are
\begin{equation}
  s_{it} = \begin{cases}
    a_i + v_{it}\,, \quad \text{if } t = 1\,,\\
    v_{it}\,, \quad \text{otherwise}
  \end{cases}
  \quad \forall i \in \mathcal{N}, t \in \mathcal{T}\,.
\end{equation}

Moreover, during real-time operations, the system may have to plan not only for predicted future customers but also for outstanding customers that were not serviced at previous time steps. Thus, in this step we count the outstanding travel requests and represent them by $\{\lambda_{ij0}\}_{ij0}$.

\subsection{Forecasting}

The second part of Algorithm \ref{alg:mpc} consists of predicting future customer demand.
Let $f$ be a forecasting model trained with historical data, $\theta_t$ a diverse set of features relevant to the model available at the time of prediction (e.g. current traffic conditions, weather, recent travel demand, etc.), and $T_{\mathrm{forward}}$ the forecasting horizon. Then, we denote $\hat{\Lambda}_{t+1,t+T_{\mathrm{forward}}} = f(\theta_t)$ as the expected demand for the forecasting horizon, where $\hat{\Lambda}_{t+1,t+T_{\mathrm{forward}}} = \{\hat{\lambda}_{ijt'}\}_{ijt'}, \forall i,j \in \mathcal{N}, t' \in [t+1,...,t+T_{\mathrm{forward}}]$. The design of techniques for forecasting customer demand is beyond the scope of this paper: we refer the reader to \cite{ZhaoTarkomaEtAl2016} for a recent review. In Section \ref{sec:numerical_experiments}, we propose a forecasting model based on neural networks and validate its performance with real customer data.

\subsection{Controller}\label{sec:controller}

The third step computes the rebalancing strategy for the planning horizon $[1,..., T]$ using the observed state and the predicted demand. To achieve this, we adapt Problem \eqref{eq:det-opt} for real-time usage by introducing several modifications to the problem formulation.

First, from the proof sketch of Theorem \ref{th:feas}, note that the ensured feasibility of \eqref{eq:det-opt} depends on being able to choose the starting vehicle positions $\mathcal{S}$. Thus, for a fixed $\mathcal{S}$ and given customer demands, \eqref{eq:det-opt} might be infeasible.
To ensure persistent feasibility of the MPC controller, we relax constraint \eqref{eq:pax-constr} by allowing the optimizer not to service certain customers. The slack variables $\mathcal{D} = \{d_{ijt}\}_{ijt}$  denote the predicted demand of customers wanting to travel from $i$ to $j$ departing at time $t$ that will remain unsatisfied.

Second, outstanding customers may be at stations currently without any available vehicles: thus, it may be infeasible to pick them up at $t=1$. To mitigate this, we let the pickup time for outstanding customers be an optimization variable (with an associated cost that is proportional to the customers' waiting time). Formally, we define  $w_{ijt}$ as the decision variable denoting the number of outstanding customers at region $i \in \mathcal{N}$ who wish to travel to region $j$ and be picked up at time $t \in \mathcal{T}$. To ensure that all outstanding customers are considered, we include the following constraint:
\begin{equation}
  \sum_{t \in \mathcal{T}} w_{ijt} = \lambda_{ij0}\,, \quad \forall i,j \in \mathcal{N}\,,
\end{equation}
where $\lambda_{ij0}$ is the number of outstanding customers wanting to go from $i$ to $j$.
Accordingly, the counterpart of Equation \eqref{eq:pax-constr} in the MPC controller is:
\begin{equation}
  x_{ijt}^p + d_{ijt} = \hat{\lambda}_{ijt} + w_{ijt}, \quad \forall i,j \in \mathcal{N}, t \in \mathcal{T}.
\end{equation}

Dropping customer demand and making outstanding customers wait are both undesirable. For a given origin-destination pair $i,j \in \mathcal{N}$, we define $c_{ijt}^w$ and $c_{ijt}^d$ as the cost associated with a wait time of $t$ time steps for an outstanding passenger, and the cost for not servicing a predicted customer demand at time $t$, respectively. 

We are now in a position to state the overall MPC optimization problem:

{\small
\begin{subequations}
\begin{align}
& \underset{\mathcal{X}^p, \mathcal{X}^r, \mathcal{W}, \mathcal{D}}{\text{minimize}} 
  & & \sum_{(ijt)} c_{ijt}^r x_{ijt}^r + c_{ijt}^w w_{ijt} + c_{ijt}^d d_{ijt},\\
  & \text{subject to} && x_{ijt}^p + d_{ijt} - w_{ijt} = \hat{\lambda}_{ijt}, \quad \forall i,j \in \mathcal{N}, t \in \mathcal{T},\label{eq:pax-constr-mpc}\\
  &
  && \sum_{j \in \mathcal{N}} x_{ijt}^p + x_{ijt}^r - x_{jit-\tau_{ij}}^p - x_{jit-\tau_{ij}}^r = s_{it}, \nonumber \\
  &
  && \quad \quad \quad \quad \forall i \in \mathcal{N}, t \in \mathcal{T}\,,\label{eq:veh-constr-mpc}\\
  &
  && \sum_{t \in \mathcal{T}} w_{ijt} = \lambda_{i,j,0}\,, \quad \forall i,j \in \mathcal{N}, \label{eq:out-constr-mpc}\\
  &
  && x_{ijt}^p, x_{ijt}^r, w_{ijt}, d_{ijt} \in \mathbb{N}, \quad \forall i,j \in \mathcal{N}, t \in \mathcal{T}\,. \label{eq:int-vars-mpc}
\end{align}
\label{eq:mpc-opt}
\end{subequations}
}
Here, \eqref{eq:pax-constr-mpc} and \eqref{eq:veh-constr-mpc} are the passenger and vehicle continuity constraints, \eqref{eq:out-constr-mpc} ensures that all outstanding passengers are served, and \eqref{eq:int-vars-mpc} limits the decision variables to nonnegative integers. 

\subsection{Discussion}

Note that the rebalancing, waiting and dropping costs are optimization parameters that should be set by the operator to reflect real-life costs. However, it is important to highlight that the relative difference between the waiting costs and the dropping costs should be carefully chosen: the optimizer will choose to drop a customer if it is cheaper than making her or him wait. Additionally, the planning horizon should be carefully chosen: 
an excessively short planning horizon may make it impossible for the optimizer to allocate rebalancing vehicles to customers stranded in remote regions, since the travel time might be longer than the planning horizon itself.
Finally, Problem \eqref{eq:mpc-opt} is {\emph{not}} totally unimodular, due to the presence of constraint \eqref{eq:out-constr-mpc}: therefore, the problem must be solved as a mixed-integer linear program (MILP). However, unlike existing MILP approaches in literature (e.g. \cite{ZhangRossiEtAl2016b,Miao2016}) the problem size of \eqref{eq:mpc-opt} does not grow with the number of vehicles -- a remarkable fact that makes this approach suitable for large fleet sizes. Indeed, in Section \ref{sec:numerical_experiments} we show that modern MILP algorithms are able to solve Problem \eqref{eq:mpc-opt} quickly for large-scale problems based on real-world data.

\section{Numerical Experiments}\label{sec:numerical_experiments}

In this section, we first present numerical experiments comparing the performance of the proposed algorithm against an existing, high-performing rebalancing heuristic \cite{PavoneSmithEtAl2012,ZhangRossiEtAl2016b}. Then, we test the sensitivity of the algorithm to the length of the forecast horizon $T_\text{forward}$. The experiments were carried on simulations based on a real-world dataset from the Chinese ridesharing company Didi Chuxing.

\subsection{Dataset}

The DiDi dataset contains all trips requested by users in the city of Hangzhou from January 1 to January 21, 2016 (approximately eight million trips).
For each trip, the dataset records the time of the request, the departure and destination locations (discretized in districts, or regions), a customer ID, a driver ID, and the price paid. Start locations are discretized into 66 districts (denoted as core districts), each identified by a hash. Destination locations are similarly discretized in 793 districts (a superset of the start locations). For the purpose of our numerical experiments, we disregarded trips that did not start and end in the core districts (approximately one million trips).

The dataset reports no geographic information about the location of the individual districts. Furthermore, no information is provided about the duration (and therefore the end time) of completed trips. However, using RideGuru \cite{ios_RideGuru:2017} we were able to estimate the travel time of each trip from the trip price; we used this estimate to reconstruct the average travel time between each pair of districts.

\subsection{Simulation environment}
We simulate the operations of an AMoD system servicing customer trips from the DiDi dataset. 
The city is modeled as a set of regions, corresponding to the districts in the DiDi dataset. Each pair of regions is connected by a road whose travel time equals the average estimated travel time computed from the DiDi dataset.

Customer requests are ``played back'': for each fulfilled transportation request in the dataset, we introduce in the simulation a customer request with the same start time, start location, and arrival location .
 If a customer request appears in a region where a vehicle is available, the customer is assigned to the vehicle and departs immediately. Otherwise, the customer waits in a queue for the next available vehicle.
Vehicles remain idle at a region until they are assigned to either a customer request or a rebalancing task. The travel time of vehicles assigned to customer requests corresponds to the travel time of the actual corresponding trip in the dataset. In contrast, the travel time of vehicles assigned to a rebalancing task is the average estimated travel time between the origin and the destination region.
The system state evolves in discrete time: every time step in the simulation corresponds to 6 seconds.

Every $\Delta t = 5$ minutes, we execute the MPC algorithm \eqref{alg:mpc}. For each region, the algorithm produced a (possibly empty) list of routes that empty vehicles should follow. The routes are then assigned to idle vehicles within each region. At the beginning of each iteration of Algorithm \eqref{alg:mpc}, the unused rebalancing tasks are deleted, and the process is repeated.

\subsection{Forecasting}\label{sec:forecasting-lstm}

Long Short-Term Memory (LSTM) neural networks \cite{HochreiterSchmidhuber1997} are a popular and effective method for forecasting time-series, and have increasingly gained attention in transportation demand forecasting (e.g. \cite{Laptev2017,Song2016,Kea2017}). 

In this paper, we built forecasting model based on an LSTM neural network following the encoder-decoder architecture \cite{Cho2014}. The model's input is a multivariate time series describing the demand for each origin and destination pair for the last $T_{\mathrm{back}}$ time steps. The model forecasts demand for each origin and destination pair for the following $T_{\mathrm{forward}}$ time steps for each region. That is, for a trained model $f$ we forecast demand as follows:
\begin{equation}
  \hat{\Lambda}_{t_0+1,t_0+T_{\mathrm{forward}}} = f(\Lambda_{t_0-T_{\mathrm{back}},t_0})\,.
\end{equation}


\subsection{Detailed results for a single day}

We simulated the day with highest activity, January 21, 2016 with 330,000 trips. We compared the performance of four different controllers:
\begin{itemize}
  \item {\emph{MPC-Perfect}}. The controller described in Algorithm \ref{alg:mpc} with a planning horizon of $T = 50$ using the exact customer demand as it appears in the dataset as a ``forecast'' for the next $T_{\mathrm{forward}} = 24$ time intervals. The controller is non-causal; however, its performance offers an upper bound on the performance of the MPC algorithm.
  \item {\emph{MPC-LSTM}}. The controller described in Algorithm \ref{alg:mpc} with a planning horizon of $T = 50$ using the model described in Section \ref{sec:forecasting-lstm} to forecast customer demand for the next $T_{\mathrm{forward}} = 24$ time intervals.
  \item {\emph{TV-Reactive}}. The controller described in Algorithm \ref{alg:mpc} with a planning horizon of $T = 50$, but with the prediction set empty. In essence, this controller is simply a time-variant planner that ``reacts'' to outstanding customer demand.
  \item {\emph{Reactive}}. The controller described in \cite{PavoneSmithEtAl2012}. The controller is based on a time-invariant model and reactive;  however, it is shown to offer superior performance compared to several state-of-the-art rebalancing algorithms in \cite{ZhangRossiEtAl2016b}.
\end{itemize}

For all scenarios, the rebalancing costs $c_{ijt}^r$ are proportional to the travel time $\tau_{ij}$. Moreover, we assume that the operator's goal is to satisfy as many customer requests as possible. Thus, the cost of dropping predicted demand, $c_{ijt}^d$, is orders of magnitude larger than the cost of rebalancing, and the cost of waiting is $c_{ijt}^w = c_{ijt}^d / T$, such that making a customer wait for the entire planning horizon is equivalent to not satisfying the request.

To find an appropriate fleet size, we solved Problem \eqref{eq:det-opt} for the selected day: we found that the minimum number of vehicles required to satisfy customer demand without any waiting  is 4206. In order to account for the effect of imperfect information, we selected a fleet size of 5000 vehicles for the ensuing simulations. The initial location of the vehicles was equally distributed among the 66 regions.
The forecasting model was trained with the first 15 days of the dataset, using a look-back horizon of $T_{\mathrm{back}} = 60$ time steps, and a forecasting horizon of $T_{\mathrm{forward}} = 24$ time steps. 

A summary of the results can be appreciated in Table \ref{tab:results}. As expected, the MPC controller with perfect information has the best performance, with a mean wait time of 3.7 seconds and a median wait time of 0. The TV-Reactive scenario has the worst performance, with a mean wait time double than that of Reactive. This is also not surprising, given that the Reactive algorithm in \cite{PavoneSmithEtAl2012} has been shown to have excellent performance, and, without any forecasts, the MPC algorithm reacts only to outstanding passengers.
Notably, however, the best performing \emph{causal} controller, MPC-LSTM, has a mean wait time {\emph{89.6\% shorter}} than that of the Reactive controller. This highlights the value of demand forecasting, even when imperfect, for operating the fleet. 

\begin{table}[ht]
\caption{Wait times for each scenario.}
\centering
  \begin{tabular}{rcccc}
   Wait time & MPC-Perfect & MPC-LSTM &  Reactive \cite{PavoneSmithEtAl2012} & TV-Reactive\\ 
  \hline
  Mean [s]  &  3.7 & 29.4 & 283.3 & 543.5 \\

  Median [s] & 0.0 & 0.0 & 72.0 & 414.0 \\
  \end{tabular}
\label{tab:results}
\end{table}

\begin{figure}[h]
    \centering
    \includegraphics[width=0.45\textwidth,trim={0 0 0 0},clip]{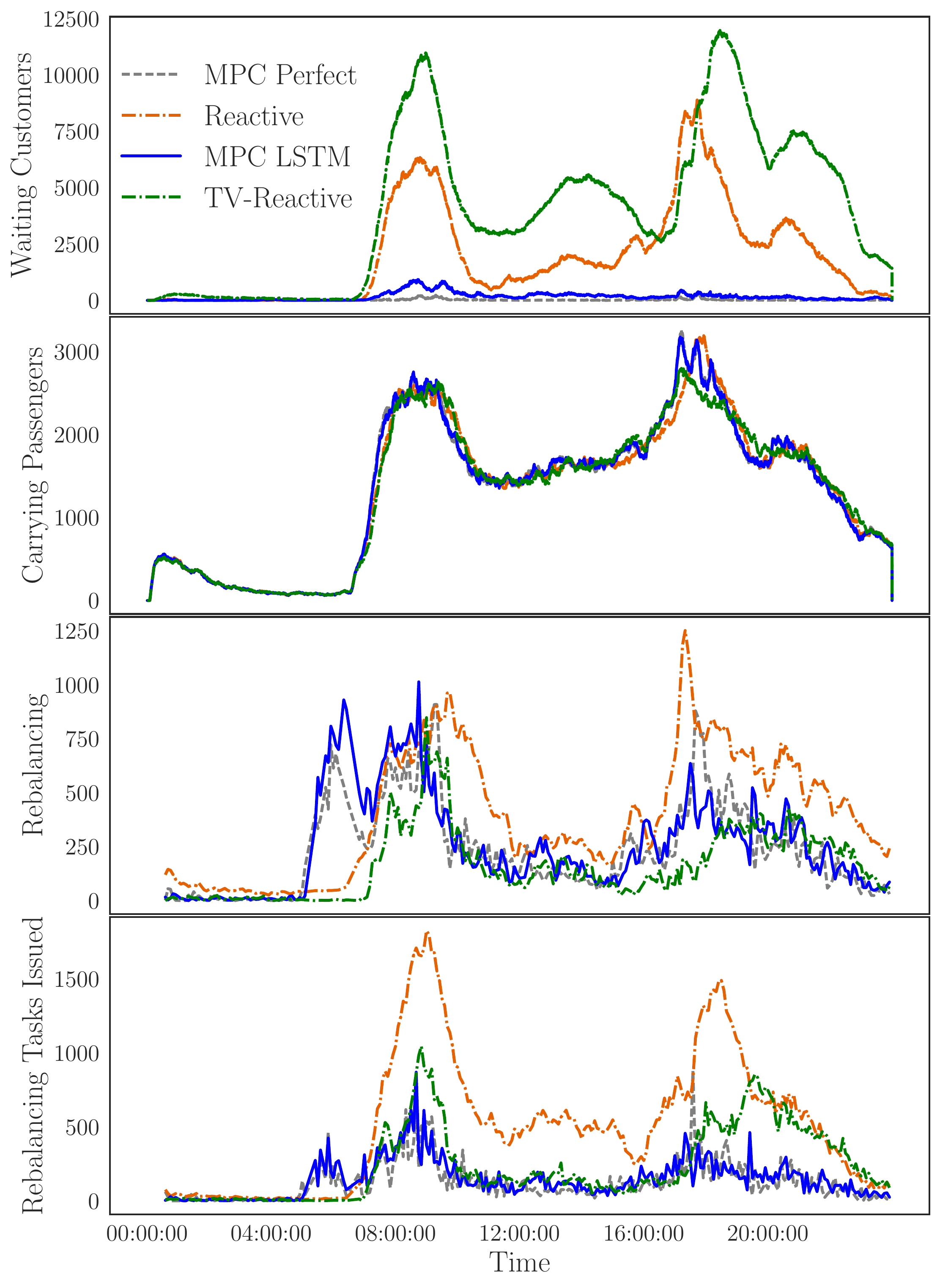}
    \caption{Results from the MPC-Perfect (gray), MPC-LSTM (blue), Reactive \cite{PavoneSmithEtAl2012} (orange), and TV-Reactive (green) scenarios at each time step. From top to bottom: Number of waiting customers; number of vehicles carrying passengers; number of vehicles rebalancing; and, number of rebalancing tasks issued.}
    \label{fig:multi-results}
  \end{figure}

More detailed results for MPC-Perfect, MPC-LSTM, and Reactive are shown in Figure \ref{fig:multi-results}. We can see the stark difference in performance from the top chart showing the number of waiting customers at any given point in time. Notably, the Reactive scenario has significantly more waiting customers at any given point in time, peaking at 8,892 in the afternoon rush. In contrast, the maximum number of waiting customers with the MPC-LSTM algorithm is 843, during the morning rush. 
Much of the performance gain can be attributed to the possibility of preemptively rebalancing enabled by the forecasts. For example, both MPC-Perfect and MPC-LSTM issue a considerable number of rebalancing tasks around 6AM, right before the morning rush, while the Reactive controller only starts rebalancing once the rush begins. 
Note, however, that not all of the difference is due to preemptive rebalancing tasks. For example, the Reactive controller issues, in total, more than 3 times as many rebalancing tasks as MPC-LSTM, but, on average, has only 37\% more vehicles rebalancing. These numbers imply that the rebalancing trips are shorter for the Reactive controller, and, thus, occur between nearby regions. However, in its attempt to keep equal availability across the city, vehicles are sent to regions where they might not be needed. 


\subsection{Comparison for different forecasting horizons}
\label{sec:forecasting-horizon}

For the next set of experiments, we tested the sensitivity of the controller to different forecasting scenarios. Specifically, we varied the forecasting forward horizon, $T_{\mathrm{forward}}$, and backward horizons, $T_{\mathrm{back}}$, to different values spanning from 3 to 48 time steps (15min to 4 hours). Thus, we trained an LSTM specifically for each $T_{\mathrm{forward}}$ and $T_{\mathrm{back}}$ combination. Note that we kept the planning horizon fixed at $T = 50$ (4 hours and 10 minutes).

\begin{figure}[h]
    \centering
    \includegraphics[width=0.45\textwidth,trim={20mm 5mm 25mm 15mm},clip]{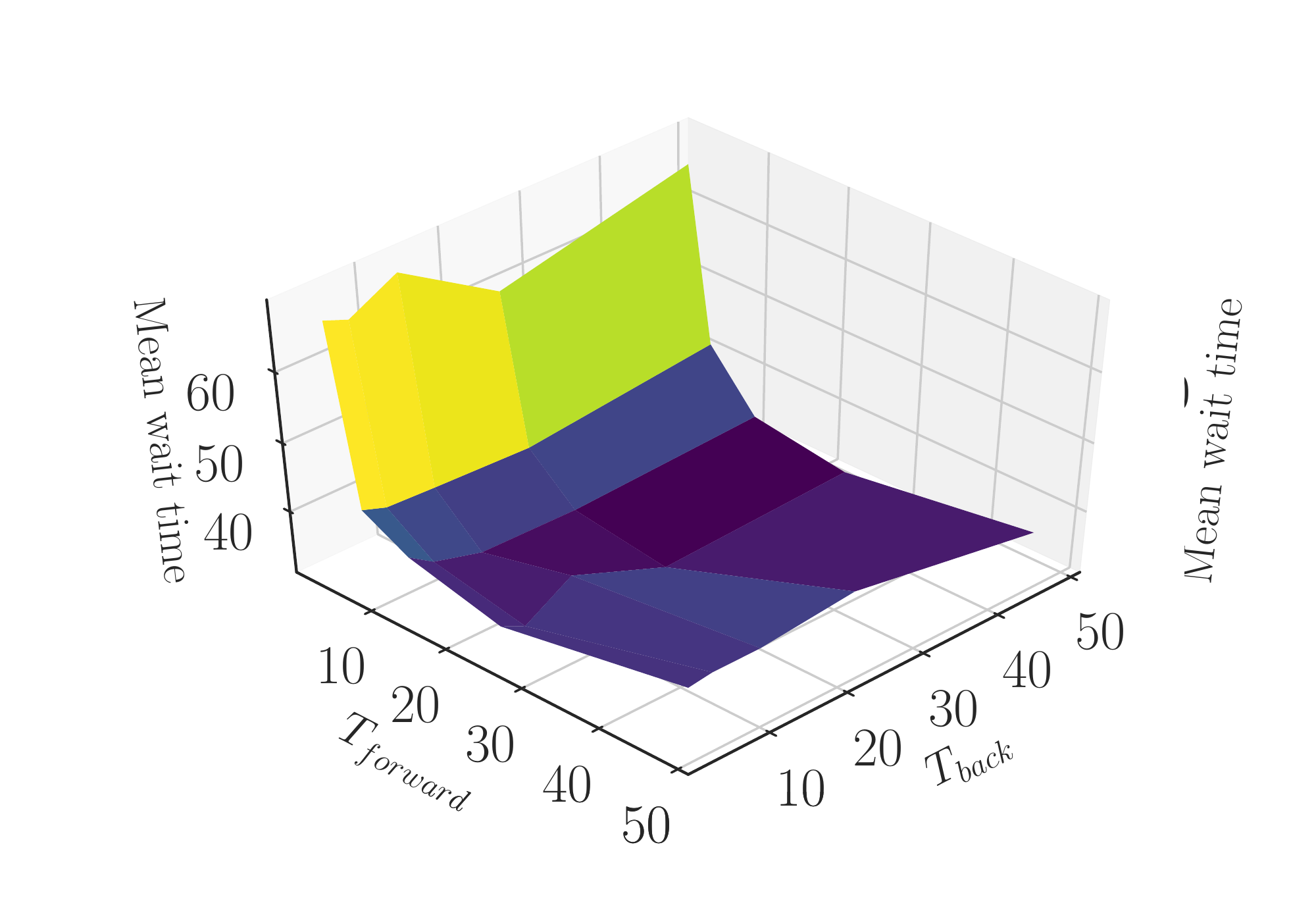}
    \caption{Mean wait times for different combinations of $T_{\mathrm{forward}}$ and $T_{\mathrm{back}}$}
    \label{fig:3dplot}
\end{figure}


Figure \ref{fig:3dplot} shows the results. On the one hand, the backward horizon $T_{\mathrm{back}}$ has little effect on the performance. This is likely due to the fact that time of the day and demand at the previous time step carry much more predictive power than the demand at even two time steps before (this was also observed empirically in \cite{Kea2017}). On the other hand, the length of $T_{\mathrm{forward}}$ follows a diminishing returns pattern, having significant, positive influence on the performance of the system at first, and leveling later. The early performance gains are likely due to the ability to foresee demand that would require long rebalancing travel times to satisfy. 
 
\subsection{Computational complexity}
In general, mixed integer linear programs require expensive computations to solve. 
However, we show in simulation that the MILP in Problem \eqref{eq:mpc-opt} can be solved efficiently on commodity hardware with state-of-the-art solvers. 
We kept track of all the MILP instances realized during the experiments in Section \ref{sec:forecasting-horizon}. The simulations were run in a PC equipped with a 3.0 GHz Intel Core i7-5960 with 64GB of RAM, and we used IBM CPLEX \cite{ios_ILOG:1987} to solve the MILP instances. Table \ref{tab:running-time} reports our results. The average time required to solve a single instance of Problem \eqref{eq:mpc-opt} was 15.1s. In 7175 instances, no instance required more than 61 seconds to solve. This suggests that the MILP can be solved in real-time for control of real transportation networks. 

\begin{table}[ht]
\caption{Optimization Running Time.}
\centering
  \begin{tabular}{rcccc}
   Samples & Mean [s] & Median [s] & STD [s] & Max [s] \\ 
  \hline
  7175 & 15.1 & 14.0 & 8.7 & 61.0 \\
  \end{tabular}
\label{tab:running-time}
\end{table}

\section{Conclusions}

In this paper, we presented a model-predictive control strategy that leveraged predicted customer demand to control the operations of an AMoD fleet. We first proposed a time-expanded network flow model for AMoD systems: the model allows us to compute the optimal rebalancing strategy and the minimum fleet size required to satisfy a given customer demand without waiting. We leverage this model to propose an MPC algorithm that relies on forecasted demand to control AMoD systems in real-time.
Numerical simulations based on real-world data show that the algorithm scales well to large systems and outperforms state-of-the-art rebalancing strategies. Collectively, our results show that the incorporating forecasted demand in the synthesis of a rebalancing algorithm can yield very significant improvements in customer satisfaction, with 89.6\% shorter customer wait times.

This paper opens several new avenues of research. 
First, it is of interest to extend this approach to incorporate other relevant and promising aspects of AMoD systems, such as including congestion, integration with the power grid, and coordination with public transit.
As the model is expanded to include these approaches, in order to preserve computational tractability, it will be necessary to explore \emph{approximate} solution techniques for mixed-integer linear programs, including rounding and randomized routing approaches.
Thus, a second line of research would devise optimization algorithms that provide constraint satisfaction guarantees and bounded suboptimality. 
Third, the MPC algorithm outperformed the state-of-the-art Reactive controller in presence of a forecast - however, its performance when no forecast is available is significantly worse than the reactive algorithm's. One possible approach to close this gap is to formulate a risk-averse MPC that takes into account uncertainty of the forecasts.
This research route will require devising forecasting models that are able to predict not only the expected demand, but also its probability distribution. 
Therefore, a fourth route is to improve existing short-term forecasting techniques such that they enhance the end-to-end performance of the control system. Models that provide well-calibrated uncertainty distributions for their forecasts are of particular interest, as are forecasting models that leverage heterogeneous sources of data, such as weather, traffic, or cellular tower records.

\bibliographystyle{IEEEtran}
\bibliography{../../../bib/main,../../../bib/ASL_papers}

\newcommand{\noopsort}[1]{} \newcommand{\printfirst}[2]{#1}
  \newcommand{\singleletter}[1]{#1} \newcommand{\switchargs}[2]{#2#1}
\begin{thebibliography}{10}
\providecommand{\url}[1]{#1}
\csname url@rmstyle\endcsname
\providecommand{\newblock}{\relax}
\providecommand{\bibinfo}[2]{#2}
\providecommand\BIBentrySTDinterwordspacing{\spaceskip=0pt\relax}
\providecommand\BIBentryALTinterwordstretchfactor{4}
\providecommand\BIBentryALTinterwordspacing{\spaceskip=\fontdimen2\font plus
\BIBentryALTinterwordstretchfactor\fontdimen3\font minus
  \fontdimen4\font\relax}
\providecommand\BIBforeignlanguage[2]{{%
\expandafter\ifx\csname l@#1\endcsname\relax
\typeout{** WARNING: IEEEtran.bst: No hyphenation pattern has been}%
\typeout{** loaded for the language `#1'. Using the pattern for}%
\typeout{** the default language instead.}%
\else
\language=\csname l@#1\endcsname
\fi
#2}}

\bibitem{Fricker2012}
C.~Fricker and N.~Gast, ``Incentives and redistribution in homogeneous
  bike-sharing systems with stations of finite capacity,'' \emph{{EURO Journal
  on Transportation and Logistics}}, 2012.

\bibitem{David2012}
G.~K. David, ``Stochastic modeling and decentralized control policies for
  large-scale vehicle sharing systems via closed queueing networks,'' Ph.D.
  dissertation, {The Ohio State University}, 2012.

\bibitem{Banerjee2016}
S.~Banerjee, D.~Freund, and T.~Lykouris. (2016) Multi-objective pricing for
  shared vehicle systems. Available at \url{https://arxiv.org/abs/1608.06819}.

\bibitem{Banerjee2015}
S.~Banerjee, R.~Johari, and C.~Riquelme, ``Pricing in ride-sharing platforms: A
  queueing-theoretic approach,'' in \emph{ACM Conference on Economics and
  Computation}, 2015.

\bibitem{PavoneSmithEtAl2012}
M.~Pavone, S.~L. Smith, E.~Frazzoli, and D.~Rus, ``Robotic load balancing for
  {Mobility-on-Demand} systems,'' \emph{{Int.\ Journal of Robotics Research}},
  vol.~31, no.~7, pp. 839--854, 2012.

\bibitem{ZhangPavone2016}
R.~Zhang and M.~Pavone, ``Control of robotic {Mobility-on-Demand} systems: A
  queueing-theoretical perspective,'' \emph{{Int.\ Journal of Robotics
  Research}}, vol.~35, no. 1-3, pp. 186--203, 2016.

\bibitem{ZhangRossiEtAl2016}
R.~Zhang, F.~Rossi, and M.~Pavone, ``Routing autonomous vehicles in congested
  transportation networks: Structural properties and coordination algorithms,''
  in \emph{{Robotics: Science and Systems}}, 2016.

\bibitem{VolkovAslamEtAl2012}
M.~Volkov, J.~Aslam, and D.~Rus, ``Markov-based redistribution policy model for
  future urban mobility networks,'' in \emph{{Proc.\ IEEE Int.\ Conf.\ on
  Intelligent Transportation Systems}}, 2012.

\bibitem{ZhangRossiEtAl2016b}
R.~Zhang, F.~Rossi, and M.~Pavone, ``Model predictive control of {Autonomous}
  {Mobility-on-Demand} systems,'' in \emph{{Proc.\ IEEE Conf.\ on Robotics and
  Automation}}, 2016.

\bibitem{Miao2016}
F.~Miao, S.~Han, S.~Lin, J.~A. Stankovic, D.~Zhang, S.~Munir, H.~Huang, T.~He,
  and G.~J. Pappas, ``Taxi dispatch with real-time sensing data in metropolitan
  areas: A receding horizon control approach,'' \emph{{IEEE Transactions on
  Automation Sciences and Engineering}}, 2016.

\bibitem{Miller2017}
J.~Miller and J.~P. How, ``Predictive positioning and quality of service
  ridesharing for campus mobility on demand systems,'' \emph{{IEEE Conf.\ on
  Robotics and Automation}}, 2017.

\bibitem{AhujaMagnantiEtAl1993}
R.~K. Ahuja, T.~L. Magnanti, and J.~B. Orlin, \emph{Network Flows: Theory,
  Algorithms and Applications}.\hskip 1em plus 0.5em minus 0.4em\relax
  {Prentice Hall}, 1993.

\bibitem{ZhaoTarkomaEtAl2016}
K.~Zhao, S.~Tarkoma, S.~Liu, and H.~Vo, ``Urban human mobility data mining: An
  overview,'' in \emph{{IEEE Int.\ Conf.\ on Big Data}}, 2016.

\bibitem{ios_RideGuru:2017}
Fare estimates, rideshare questions \& answers. RideGuru. {https://ride.guru/}.

\bibitem{HochreiterSchmidhuber1997}
S.~Hochreiter and J.~Schmidhuber, ``Long short-term memory,'' \emph{{Neural
  Computation}}, 1997.

\bibitem{Laptev2017}
N.~Laptev, J.~Yosinski, L.~E. Li, and S.~Smyl, ``Time-series extreme event
  forecasting with neural networks at uber,'' in \emph{{Int.\ Conf.\ on Machine
  Learning}}, 2017.

\bibitem{Song2016}
X.~Song, H.~Kanasugi, and R.~Shibasaki, ``Deeptransport: Prediction and
  simulation of human mobility and transportation mode at a citywide level.''
  \emph{{International Joint Conference on Artificial Intelligence}}, 2016.

\bibitem{Kea2017}
J.~Kea, H.~Zhengb, H.~Yanga, and X.~M. Chenb. (2017) Short-term forecasting of
  passenger demand under on-demand ride services: A spatio-temporal deep
  learning approach. Available at \url{https://arxiv.org/abs/1706.06279}.

\bibitem{Cho2014}
K.~Cho, B.~van Merrienboer, C.~Gulcehre, D.~Bahdanau, F.~Bougares, H.~Schwenk,
  and Y.~Bengio. (2014) Learning phrase representations using rnn
  encoder-decoder for statistical machine translation. Available at
  \url{https://arxiv.org/abs/1409.1259}.

\bibitem{ios_ILOG:1987}
\emph{ILOG CPLEX User's guide}, IBM ILOG, 1987.

\end{thebibliography}
\end{document}